\def\paperID{0006}
\def\confName{ICCV - NeVI Workshop}
\def\confYear{2025}

\def\paperTitle{Quantifying Accuracy of an Event-Based Star Tracker via Earth's Rotation}

\def\authorBlock{
    Dennis Melamed\qquad 
    Connor Hashemi \qquad 
    Scott McCloskey \\
    Kitware \\
    {\tt\small dennis.melamed@kitware.com}
}

\newif\ifreview 
\newif\ifarxiv \newcommand{\arxiv}{\arxivtrue}
\newif\ifcamera 
\newif\ifrebuttal

\newcommand{\ebcincelestial}[1]{^CR_e (#1)}
\newcommand{\ebcinterrestrialestim}{^T\hat{R}_e}
\newcommand{\ebcinterrestrialreal}{^TR_e}
\newcommand{\earthincelestial}[1]{^CR_T (#1)}

\newcommand{\acrosserr}{18.47}
\newcommand{\abouterr}{74.84}

\newcommand{\codeurl}{\href{https://gitlab.kitware.com/nest-public/telescope_accuracy_quantification}{gitlab.kitware.com/nest-public/telescope\_accuracy\_quantification}}
\arxiv
\pdfoutput=1
\documentclass[10pt,twocolumn,letterpaper]{article}
\ifreview \usepackage[review]{cvpr} \fi
\ifarxiv \usepackage[pagenumbers]{cvpr} \fi
\ifrebuttal \usepackage[rebuttal]{cvpr} \fi
\ifcamera \usepackage{cvpr} \fi

\usepackage{graphicx}	
\usepackage{amsmath}	
\usepackage{amssymb}	
\usepackage{booktabs}
\usepackage{times}
\usepackage{epsfig}
\usepackage{caption}
\usepackage{float}
\usepackage{placeins}
\usepackage{color, colortbl}
\usepackage{stfloats}
\usepackage{enumitem}
\usepackage{tabularx}
\usepackage{xstring}
\usepackage{multirow}
\usepackage{xspace}
\usepackage{url}
\usepackage{xcolor}
\usepackage[hang,flushmargin]{footmisc}

\ifcamera \usepackage[accsupp]{axessibility} \fi

% ...

\ifarxiv  \fi

\newcommand{\R}[1]{{%
    \textbf{%
        \ifstrequal{#1}{1}{\textcolor{red}{R#1}}{%
        \ifstrequal{#1}{2}{\textcolor{blue}{R#1}}{%
        \ifstrequal{#1}{3}{\textcolor{magenta}{R#1}}{%
        \ifstrequal{#1}{4}{\textcolor{teal}{R#1}}{%
                           \textcolor{cyan}{R#1}%
        }}}}%
    }%
}}

\usepackage{xr-hyper}

\makeatletter
\newcommand*{\addFileDependency}[1]{
  \typeout{(#1)}
  \@addtofilelist{#1}
  \IfFileExists{#1}{}{\typeout{No file #1.}}
}

\makeatother
\newcommand*{\myexternaldocument}[1]{
    \externaldocument{#1}
    \addFileDependency{#1.tex}
    \addFileDependency{#1.aux}
}

\definecolor{cvprblue}{rgb}{0.21,0.49,0.74}
\usepackage[pagebackref,breaklinks,colorlinks,allcolors=cvprblue]{hyperref}
\usepackage[capitalize]{cleveref}
\crefname{section}{Sec.}{Secs.}
\crefname{table}{Table}{Tables}
\crefname{figure}{Fig.}{Figs.}

\ifarxiv \crefname{appendix}{App.}{Apps.}
\else \crefname{appendix}{Suppl.}{Suppls.} \fi

\usepackage[accsupp]{axessibility}  

\unless\ifarxiv \myexternaldocument{_supplementary} \fi

\ifcamera
\newcommand{\codeloc}{\codeurl{}}
\fi
\ifreview
\newcommand{\codeloc}{\codetbd{}}
\fi
\ifarxiv
\newcommand{\codeloc}{\codeurl{}}
\fi
\begin{document}
\title{\paperTitle}
\author{\authorBlock}
\maketitle

\begin{abstract}
Event-based cameras (EBCs) are a promising new technology for star tracking-based attitude determination, but prior studies have struggled to determine accurate ground truth for real data.
We analyze the accuracy of an EBC star tracking system utilizing the Earth's motion as the ground truth for comparison.
The Earth rotates in a regular way with very small irregularities which are measured to the level of milli-arcseconds. 
By keeping an event camera static and pointing it through a ground-based telescope at the night sky, we create a system where the only camera motion in the celestial reference frame is that induced by the Earth's rotation. 
The resulting event stream is processed to generate estimates of orientation which we compare to the International Earth Rotation and Reference System (IERS) measured orientation of the Earth.
The event camera system is able to achieve a root mean squared across error of \acrosserr{} arcseconds and an about error of \abouterr{} arcseconds.
Combined with the other benefits of event cameras over framing sensors (reduced computation due to sparser data streams, higher dynamic range, lower energy consumption, faster update rates), this level of accuracy suggests the utility of event cameras for low-cost and low-latency star tracking. 
We provide all code and data used to generate our results\footnote{\codeloc{}}.

\end{abstract}
\section{Introduction}
\label{sec:intro}
\begin{figure*}
    \centering
    \includegraphics[width=\linewidth]{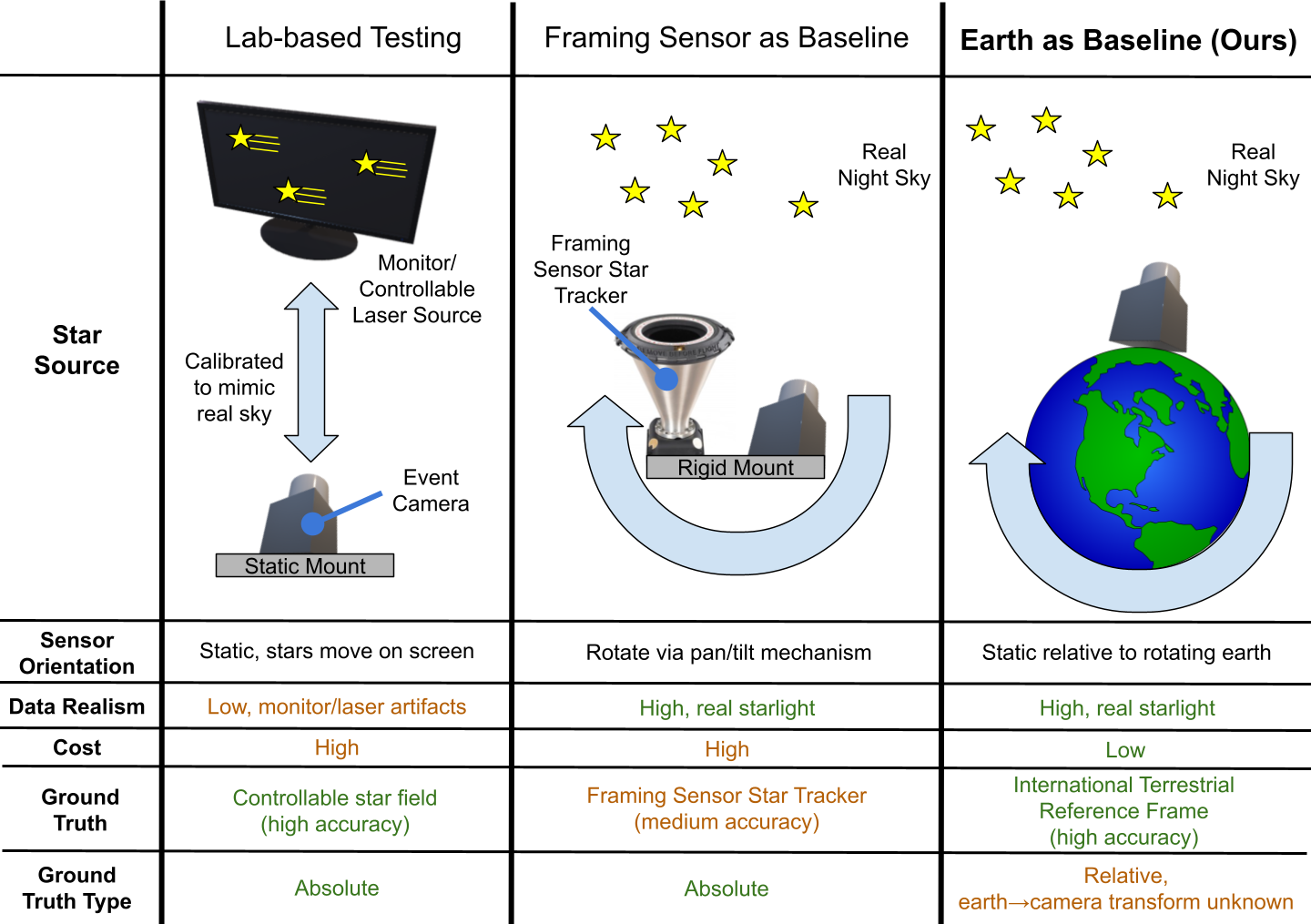}
    \caption{We measure the accuracy of an event camera-based star tracker using the rotation of the earth as ground truth (right). 
    Prior methods to test star trackers have either used controllable artificial sources of star light (monitors/lasers - left) to derive ground truth, or used a framing sensor star tracker as a source of pseudo-ground truth (center).
    While both of these approaches have benefits, they tend to require expensive equipment (high FPS monitors, laser systems, or framing sensor star trackers) and be complex to calibrate. 
    The approach presented here requires only the event camera under test and utilizes the freely available sub-milliarcsecond accuracy estimates of the earth's orientation from the International Earth Rotation and Reference Systems Service (IERS). 
    In addition our approach utilizes real starlight, avoiding the artifacts of artificial sources like frame refresh.
    Only utilizing the sensor under test minimizes issues surrounding differential flexure when performing ground-based assessment using another sensor (e.g. framing sensor star tracker) as a reference. 
    The ground truth provided by IERS is `relative,' as the static transform between the event camera frame (test estimates) and the earth frame (ground truth estimates) is unknown.
    We show a mechanism for generating an estimate of this transform.}
    \label{fig:teaser}
\end{figure*}

Star tracking is an attitude determination method, frequently utilized in space settings, which relies on the fixed location of stars in the night sky.
Star locations are extracted from images taken by cameras pointing at deep space, and a database lookup of these star positions relative to each other is used to determine the attitude of the imaging device. 
Accurate attitude estimation is critical for missions which require precise pointing of science and communication instruments, making star trackers a common component of satellites. 

Event-based cameras (EBCs) are an emerging technology with applications in robotics and motion detection.
They consist of an array of pixels which trigger asynchronously when their illumination levels change by a threshold amount.
The output of an event camera is a series of events ($x,y,p,t$) which encode which pixel measured an illumination change ($x$,$y$ location), a microsecond resolution timestamp of when the change occurred ($t$), and if the change was an increase or decrease in illumination ($p$). 
An example of stars as imaged by an event sensor is shown in \Cref{fig:experiment_config}(a).
Due to an EBC's asynchronous reaction to illumination change, event data is significantly sparser than full framing sensor images while still being able to capture changes that normally require high frame rate (\textgreater1000Hz) imaging. 
This is especially true when imaging stars, as events are only generated from pixels illuminated by the stars and data bandwidth is not used transmitting the dark sky regions of the image.
Event sparsity has the additional benefit of reducing the amount of power consumed by the camera. 
Only background maintenance power is used if there are no intensity changes to transmit, helping event cameras stay within the highly limited power envelope available in space applications.
Because event pixels report changes asynchronously as they happen, event sensors do not have a sensor-wide exposure mechanism and thus have a higher dynamic range than their framing sensor counterparts. 
Increased dynamic range allows for continued tracking of dim stars when undesirable light from the moon, sun, or earth enters the optical system. 
Robustness to stray light would allow the large exclusion hoods used on framing sensor star trackers (see center column of \Cref{fig:teaser}), used to limit stray light, to be reduced in size or removed entirely. 
EBCs are thus an excellent choice for space applications and star tracking in particular\cite{mchargFalconNeuroSpacebased2024}.

EBCs have been utilized previously for star tracking-based attitude determination, but prior studies \cite{reedEBSEKFAccurateHigh2025, bagchi_event-based_2020, chin_star_2019, Ng2022, latif_high_2023} have struggled to determine accurate ground truth for real data due to EBCs' unique imaging modality which makes applying conventional techniques difficult.
Prior approaches to estimating star tracker accuracy have pointed the tracker at a screen displaying a simulated star field \cite{bagchi_event-based_2020, chin_star_2019, Ng2022, latif_high_2023}, as on the left side of \Cref{fig:teaser}.
These approaches do not necessarily work for EBCs because event cameras are temporally sensitive enough to detect monitor artifacts, like refresh rates, which are not present in real star fields.
Comparing event camera to framing star trackers as ground truth (\Cref{fig:teaser}, center) presents its own issues, as the framing sensors have accuracy limitations themselves\cite{reedEBSEKFAccurateHigh2025}.
The cost of purchasing framing sensor star trackers for comparison can also be prohibitive, with engineering models costing many thousands of dollars. 
These difficulties prevent the rigorous evaluation necessary for inclusion in space systems from being performed for EBC-based star trackers.

In this work, we measure the accuracy of an event camera-based star tracking system (EBS-EKF\cite{reedEBSEKFAccurateHigh2025}) against highly accurate estimates of the Earth's orientation in celestial frame (\Cref{fig:teaser}, right). 
Earth's orientation is measured from a fusion of data from Global Navigation Satellite Systems (GNSS), Satellite and Lunar Laser Ranging (SLLR), Very Long Baseline Interferometry (VLBI), and Doppler Orbitography and Radiopositioning Integrated by Satellite (DORIS)\cite{wooden_explanatory_2004} which are combined and published by the International Earth Rotation and Reference Systems Service (IERS)\cite{luzum_iers_2014}.
These estimates are expressed as the orientation of the International Terrestrial Reference Frame (ITRF) in the International Celestial Reference Frame (ICRF). 
EBS-EKF outputs camera attitude estimates in the ICRF, so to quantify accuracy we calculate error between the ITRF and EBS-EKF estimates. 
As a result of the proposed method, we are able to obtain a precise evaluation of EBS-EKF accuracy with minimal calibration and with minimal sources of error.

\section{Related Work}
\label{sec:related}

Recent work has developed a multitude of ways to estimate sensor attitude using event streams. 
Initial approaches integrated events into synchronous frames which mimic the output of framing cameras and applied standard stellar motion estimation techniques \cite{chin_star_2019}.
A later approach used the well-defined spatiotemporal streaks generated in the event space to track via the Hough Transform \cite{bagchi_event-based_2020}.
An image-space 2D Kalman Filter was the first to use events as they asynchronously arrived to update the sensor attitude estimate \cite{Ng2022,latif_high_2023}, and a more recent work titled EBS-EKF utilized more accurate event-generation signal models and a 3D Kalman filter to achieve state estimation for high rotational speeds and output frequencies \cite{reedEBSEKFAccurateHigh2025}.
EBS-EKF utilizes the visibly non-Gaussian nature of the event distribution around each star to improve star centroiding, leading to increases in accuracy.

A remaining difficulty in applying EBCs to star tracking has been precise performance validation in realistic scenarios. 
Ground truth orientations for an EBC star tracker have been generated using one of two approaches: (1) simulating star fields with ground truth attitude values \cite{chin_star_2019,bagchi_event-based_2020,Ng2022,latif_high_2023,liuNovelApproachLaboratory2010,xiong_high-accuracy_2015} or (2) using off-the-shelf star trackers with a framing sensor to generate pseudo-ground truth estimates \cite{reedEBSEKFAccurateHigh2025}. 
Each of these approaches has limitations. 
Simulated data can be produced by imaging a star field displayed on a monitor corresponding to a known camera orientation. 
This allows for highly accurate ground truth and works well for conventional framing cameras \cite{nardinoMainTechnicalFeatures2021}. 
However, the high frequency change detection of an event camera causes artifacts due to the refresh rate of the monitor and unrealistic noise patterns which are not present in real star data.
More complex star simulation systems using laboratory laser systems exist which do not have monitor issues \cite{liuNovelApproachLaboratory2010}, but are complex to calibrate and expensive to purchase.
Using a framing sensor star tracker\cite{reedEBSEKFAccurateHigh2025} or a gyroscope\cite{xiong_high-accuracy_2015} as a baseline introduces possible inaccuracies into the ground truth due to the other sensor's own estimation error and misalignment/flexure between the framing and event-based trackers.

Earth rotates in a highly predictable way due to its regular daily cycle, polar motion, nutation, and precession.
The specific orientation at any given time is known to a high degree of accuracy, on the order of milli-arcseconds \cite{kurEvaluationSelectedShortterm2022}, and is formalized as the International Terrestrial Reference Frame (ITRF).
The ITRF is estimated in the International Celestial Coordinate Frame (ICRF) by the International Earth Rotation and Reference Systems Service (IERS) using multiple instruments and measurement methods\cite{wooden_explanatory_2004, luzum_iers_2014}.
The accuracy of these estimates is high enough that errors are negligible for assessing smaller star trackers with multi-arcsecond root mean square errors (RMSEs).
Earth's rotation thus provides a useful ground truth to help quantify star tracker accuracy and has been used to perform pre-launch star tracker calibration\cite{hanLowbudgetCubeSatStar2022}.

\section{Method}
\label{sec:method}

\begin{figure*}
    \centering
    \includegraphics[width=\linewidth]{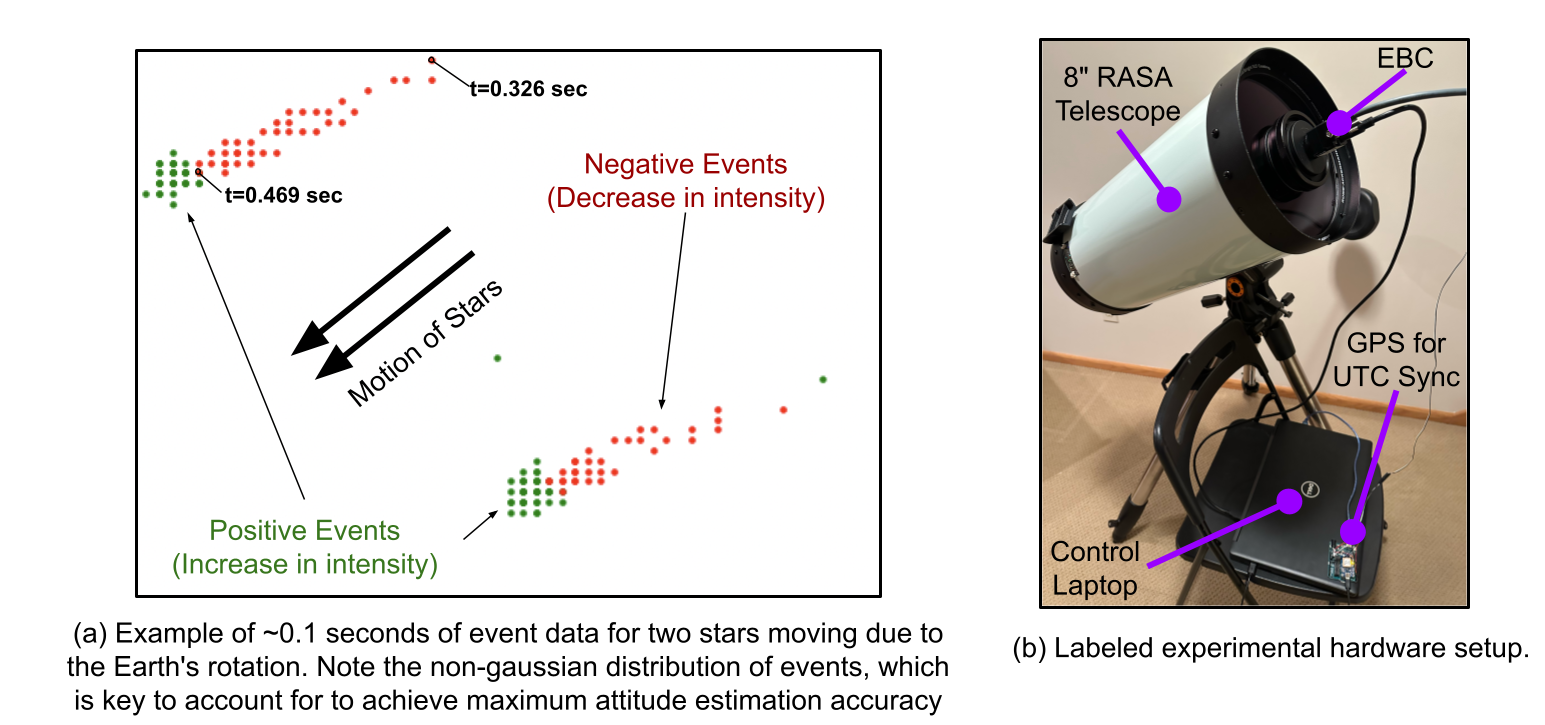}
    \caption{Experimental Configuration}
    \label{fig:experiment_config}
\end{figure*}

\subsection{Optical System}

Our optical system, shown in \Cref{fig:experiment_config}(b), consists of a Metavision EVK4-HD event-based camera mounted to an 8 inch aperture Rowe-Ackerman Astrograph (RASA) telescope with a 400mm focal length.  
The telescope includes a light-pollution filter blocking narrow bands of light wavelengths common to artificial sources such as the 440, 545, 580, and 630 nanometer regions.
The complete system provides a 0.89x0.505 degree FOV for the 1280x720 pixel event camera image sensor.
Each pixel covers 2.52 arcseconds and has a dynamic range of over 120dB. 

\subsection{EBS-EKF}

We select EBS-EKF \cite{reedEBSEKFAccurateHigh2025} as our EBC star tracking method due to its out-performance of prior methods.
The two key features which differentiate EBS-EKF from other event-based star tracking algorithms are a 3-dimensional Extended Kalman Filter (EKF) and a magnitude-dependent star centroiding method. 
The most significant improvement in performance comes from the filter, which tracks camera attitude in full $SO(3)$ space and rigorously handles the non-linear effects of roll about the optical axis. 
Filtering in the full $SO(3)$ space reduces errors by hundreds of arcseconds compared to the next best method. 

EBS-EKF's star centroiding corrects the magnitude dependent offset between the centroid of positive events associated with a star and the star's true image plane location.
Brighter stars tend to generate events further from the true star location.
The correction is particularly important in correcting roll errors, usually reducing error by an average of 10-30 arcseconds with occasional larger decreases across the tracks reported. 
Computing the magnitude-offset curve requires calibration data of true star locations relative to their events.
We are unable to co-locate a framing sensor with the event camera on our telescope so cannot determine the true star locations. 
Thus, we do not apply the magnitude-dependent offset in this work.

In addition to EBS-EKF, we also measure the accuracy of a baseline plate-solving method which we call `astrometry' in our results.
Astrometry collects positive events into batch frames and runs \texttt{astrometry.net} \cite{hogg_automated_2008} plate-solving algorithms on these frames to determine attitude, similar to the approach used by conventional framing star trackers.


\subsection{Ground Truth Determination}

Ground truth is calculated via \texttt{skyfield} \cite{rhodesSkyfieldHighPrecision2019}, a python API for access to the IERS Earth Orientation Parameters (EOPs). 
\texttt{skyfield} EOPs are calculated internally using IERS Bulletin A data, a fusion of multiple high-accuracy measurements of the orientation of the ITRF coordinate frame in the ICRF frame.
\texttt{skyfield} EOPs are indexed by UTC timestamps. 
To determine the UTC timestamps of the event-based orientation estimates, we utilize the EVK4-HD's ability to record voltage rising/falling edges from an external input line.
A GPS unit (MTK3333) with a pulse-per-second (PPS) output is used to generate a voltage pulse sent to the EVK4-HD's external input line.
The UTC timestamp corresponding to the pulse is then read as an NMEA sentence on the control computer and stored to disk. 
At runtime, the rising edge of the pulse is stored in the event stream as an external event along with a timestamp in the event timescale. 
Having both the UTC timestamp of the rising edge and the event-timescale timestamp, we produce a linear interpolation mapping between the two allowing event-based attitude estimates to have a precise UTC time assigned.
\texttt{skyfield} is then used to look up the IERS Earth orientations for the event timestamps at which the event camera produces an attitude estimate.


At this point, for a given time $t$ we have determined the Earth's orientation (ITRF) in the ICRF frame ($\earthincelestial{t}$), and used EBS-EKF to estimate the event camera's orientation in the ICRF frame ($\ebcincelestial{t}$).
The true EBC orientation in the ITRF frame ($\ebcinterrestrialreal{}$) is an unknown, but constant value in the case where the camera does not move.
The initial EBC attitude estimate 
$\ebcincelestial{0}$ is assumed to be correct, and is back-transformed through the ICRF frame to provide an estimate of the star tracker coordinate frame relative to the Earth's ($\ebcinterrestrialestim{}$ ):
\begin{align}
    \ebcinterrestrialestim{} = (\earthincelestial{0}^{-1}) \ebcincelestial{0}.
\end{align}

$\ebcinterrestrialestim{}$ is then right-multiplied to each IERS Earth attitude $\earthincelestial{t}$ to create a `virtual telescope.' 
The virtual telescope's orientation is the IERS estimate of the EBC's orientation at timestamp $t$, which can be directly compared to the EBS-EKF estimate at timestamp $t$. 
This approach provides a relative understanding of orientation error, anchored on the assumption that the first estimate is accurate. 

\section{Results \& Analysis}
\label{sec:results}
\subsection{Data}

We collected an event stream of the night sky using our event camera and telescope system in a Bortle class 5 area. 
The telescope was kept static relative to the Earth using a heavy equatorial mount and tripod (\Cref{fig:experiment_config}(b)).  
Data was collected for one hour, enough time for the earth to rotate by approximately 15 degrees. 

\subsection{Metrics}
To compare estimates of star tracker orientation to the IERS ground truth, we compute mean angular error in arcseconds along the right ascension (RA), declination (Dec), and roll axes of the ICRF.  
Star tracker accuracy is usually measured in terms of \textbf{across} and \textbf{about} accuracy, indicating the star tracker's performance for movements sweeping \textbf{across} the image plane (RA/dec) and \textbf{about} the optical axis (roll).

\subsection{Analysis}

\begin{figure*}
    \centering
    
    \includegraphics[width=\linewidth]{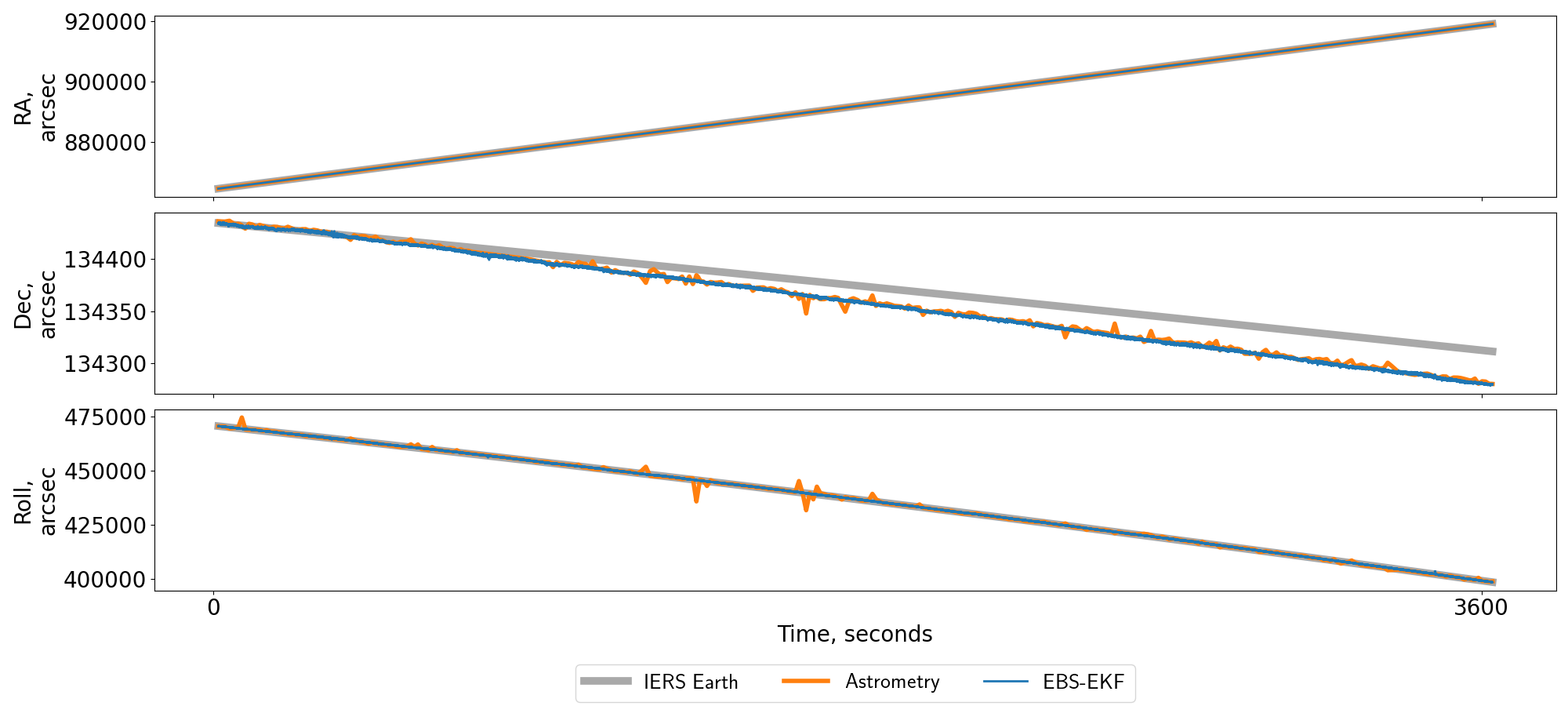}
    \caption{EBS-EKF and event-based Astrometry estimates of attitude versus known Earth rotation (IERS) rotated into the same frame.}
    \label{fig:ebs_results_fig}
\end{figure*}

\begin{figure*}
    \centering
    \hspace*{-1cm}
    \includegraphics[width=1.15\linewidth]{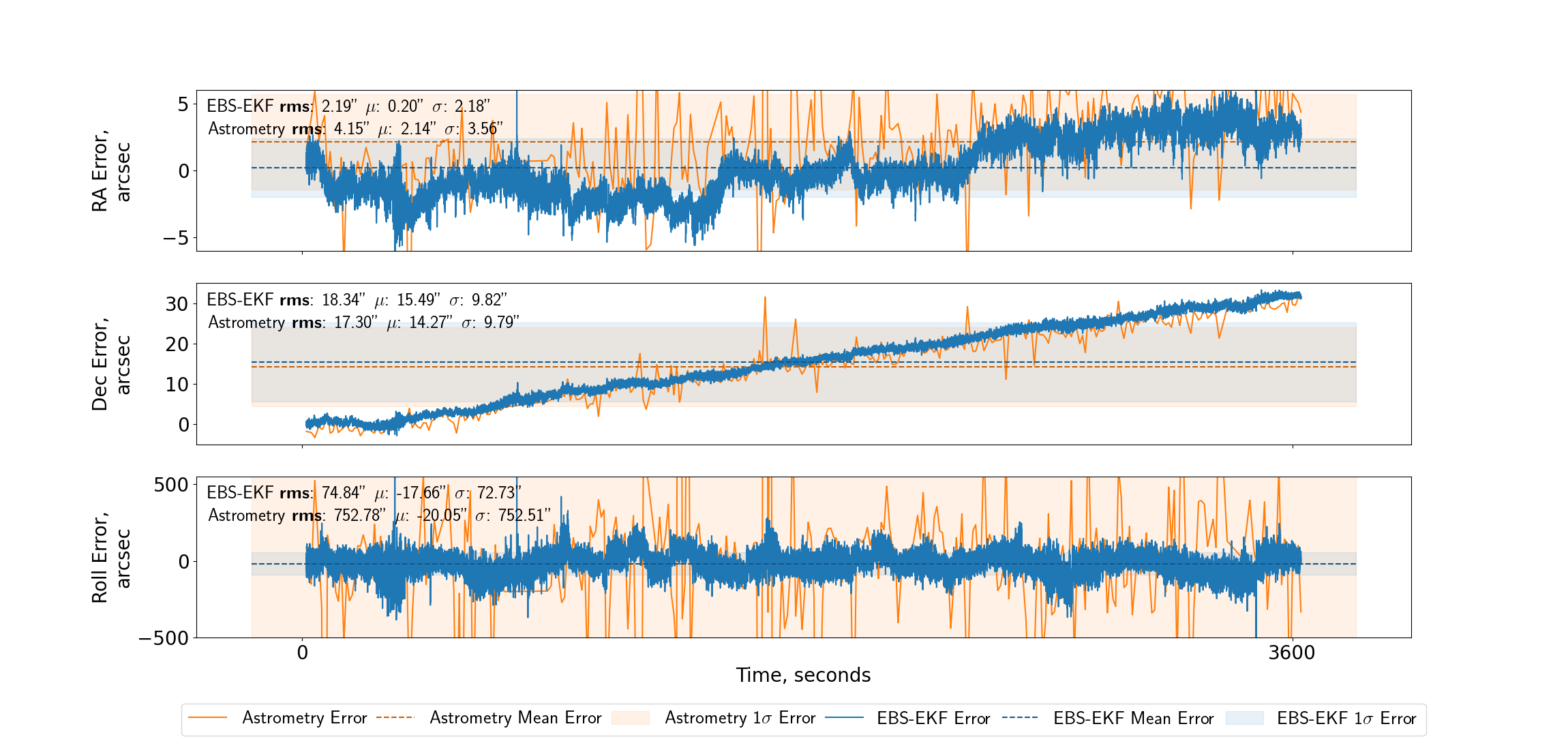}
    \caption{EBS-EKF and Astrometry error in arcseconds relative to IERS earth orientation estimates on a one hour static data collection. Astrometry errors frequently extend outside of the plot.}
    \label{fig:ebs_results_err_fig}
\end{figure*}

We set EBS-EKF to generate an estimate of the EBC's attitude at 20Hz.
We were also able to generate estimates at higher frequencies (\textgreater 100 Hz) with an approximately 2-3x reduction in accuracy.
As a point of reference the RocketLabs ST-16 framing sensor star tracker, which is commercially available with units on orbit, is able to provide 2-5 Hz attitude updates.
Our comparison method, astrometry, was able to successfully plate solve for most of the hour from $1/6$ second batches of positive events at 6Hz.

The EBS-EKF and astrometry attitude estimates for the collected hour of data, compared with the IERS Earth known attitudes, are plotted in \Cref{fig:ebs_results_fig}.
Our system achieved a root mean squared error (RMSE) of \acrosserr{} arcseconds of across error and \abouterr{} arcseconds of about error.
Error in arcseconds for each axis is shown in \Cref{fig:ebs_results_err_fig}. 
The RocketLabs ST-16 tracker reports 5 arcseconds of RMS across error and 50 arcseconds of RMS about error via an unknown evaluation method. 
We note that the optics of the ST-16 are different from our telescope, making a direct and fair comparison difficult.

\begin{figure}
    \centering
    \includegraphics[width=\linewidth]{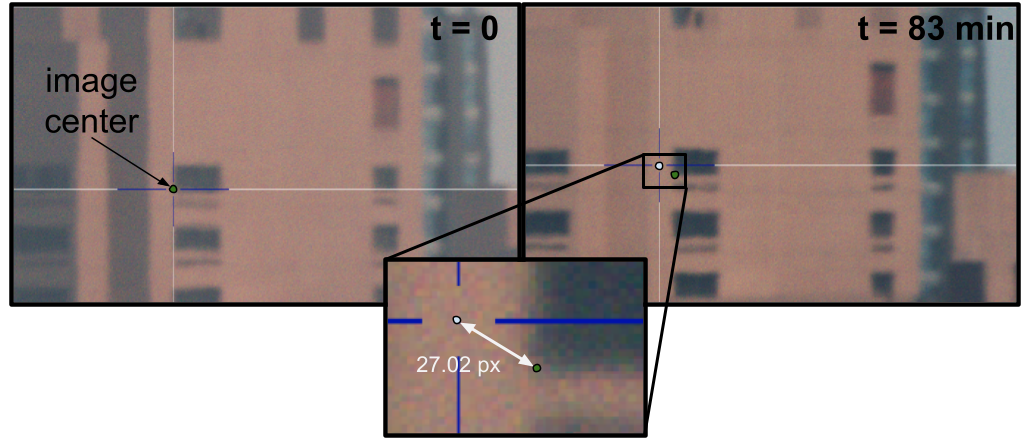}
    \caption{Image center of framing camera movement when attached to our telescope \& mount over the course of an hour. 83 minutes is enough to induce a movement of 27.02 pixels, which corresponds to a drift of 68.00 arcseconds or 49.15 arcseconds/hour.}
    \label{fig:mount_drift_fig}
\end{figure}

The majority of the across error comes from a steady drift in the declination. 
Both estimation methods (EBS-EKF and astrometry) appear to be estimating the correct trend in declination as compared to the Earth estimates (\Cref{fig:ebs_results_fig}, center), but overestimate the rate of change by approximately 30 arcseconds/hour.
EBS-EKF and astrometry showing the same trend indicates that the drift is not caused by the EBS-EKF tracking algorithm.
We determine the source of this error to be a slow movement in our equatorial mount's declination axis. 
To verify this effect we collect framing images using our telescope of a building approximately 5.8 kilometers away. 
We aim the telescope to place the center of the image at a stable reference point, the corner of the window in \Cref{fig:mount_drift_fig}. 
After 83 minutes, we compare the new image center to the image location of the window corner. 
We measure a drift in the location of the image center of 27.02 pixels, which for the focal length of our optical system corresponds to a 68.00 arcsecond shift or a 49.15 arcsecond/hour drift. 
The direction of motion corresponds to a gravity-induced rotation in our mount's declination axis, and the drift's magnitude matches what we see in the event-camera-based collection. 
\paragraph{Focal Length Analysis}

\begin{figure}
    \centering
    \includegraphics[width=1\linewidth]{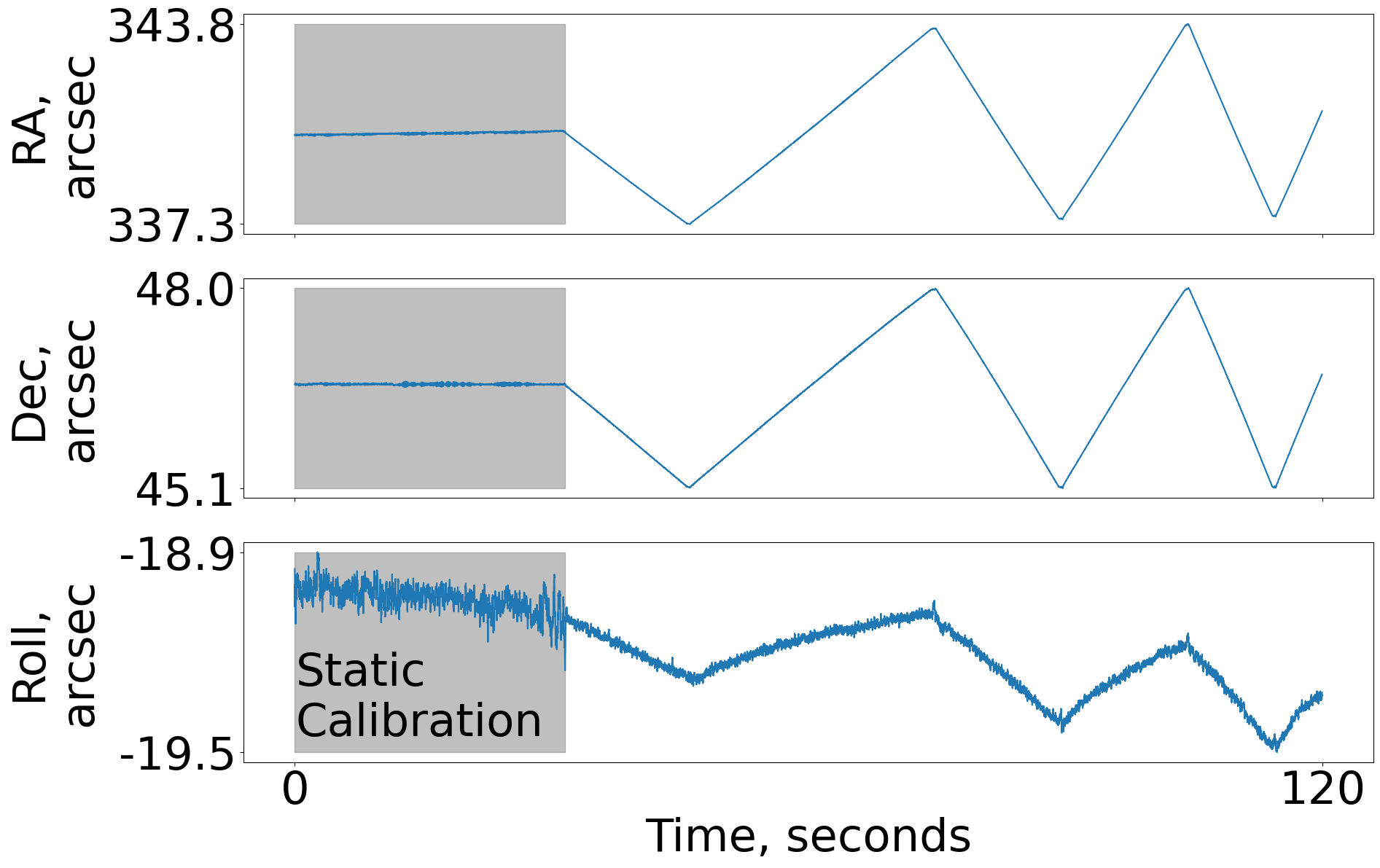}
    \caption{Right Ascension (RA), Declination (Dec), and Roll estimated by EBS-EKF on the first 2 minutes of Velocity Track 3 from the original EBS-EKF work. 
    Note the approximately 30 second static calibration period, during which the sensor is not moved by the pan/tilt unit. 
    During this period EBS-EKF is still able to track the sensor's orientation, indicating that enough events are being generated by the motion of the Earth to support tracking. 
    }
    \label{fig:vel_sweep_3_res}
\end{figure}

In this experiment we utilize a telescope with a long (400mm) focal length to make the rotation of the Earth evident in the event stream.
To better understand the necessity of using a telescope, we analyze some of the original published EBS-EKF data to determine what ranges of focal lengths would still allow tracking the Earth's rotation. 
Note that we are only analyzing the effect that the focal length has on the field of view and therefore on the apparent speed of the stars across the sensor. 
For this discussion we are ignoring other optical effects (e.g. increased aperture size and therefore increase light gathering potential) as those are more suited for selection during star tracker design.

The original EBS-EKF work used a 35mm lens on the same EVK4 event sensor as we utilize with our telescope.
\Cref{fig:vel_sweep_3_res} shows EBS-EKF tracking results for Velocity Sweep 3 from the EBS-EKF dataset.
Velocity Sweep 3 contains a static calibration period, as indicated by the gray box in \Cref{fig:vel_sweep_3_res}, where the sensor is not moving. 
The only motion present during this time is that of the Earth's rotation, which suggests a 35mm focal length is long enough for the Earth's rotation (15 \textdegree/hour) to trigger trackable events.
EBS-EKF also reports being able to accurately track while undergoing angular velocities of 1.8\textdegree/sec (Velocity Sweeps 1-5).

We can convert both of these angular velocities into image plane star velocities in pixels/second ($p$) using a pinhole camera model:
\begin{align}
    p = \frac{f * tan(s)}{x},
    \label{eq:pinhole}
\end{align}
where $f$ is the focal length in meters ($f=0.035$), $x$ is the pixel size in meters for the EVK4 ($x = 4.86*10^{-6}$), and $s$ is the angular velocity in \textdegree/sec.
For $s=15^{\circ}/hour$, the image plane pixel speed for the 35mm lens is 0.52 px/sec. 
For $s=1.8^{\circ}/sec$, the image plane pixel speed is 226.32 px/sec.
We can rearrange \Cref{eq:pinhole} to calculate the focal length required for the Earth's rotation to produce each of these image plane speeds, resulting in a range of focal lengths between \textbf{35mm} and \textbf{15.125m} which should reliably generate EBS-EKF trackable events.
This estimate is conservative, as EBS-EKF supports angular velocities of up to 7 \textdegree/sec at an unknown accuracy level. 
Additionally, focal lengths shorter than 35mm may be able to produce trackable events, but data is unavailable for these focal lengths or their image plane speed equivalents.

\subsection{Limitations}
The original EBS-EKF work introduced a velocity and magnitude dependent centroiding offset to improve attitude estimation accuracy.
The relevant calibration curve is optics-dependent and we do not recalibrate the curve for our system.
The sharp jumps in the roll error in the bottom plot of \Cref{fig:ebs_results_err_fig} map to brighter stars entering and leaving the field of view, introducing or removing a constant centroiding error into the estimation framework.
Calibration of the centroid offsetting would likely reduce the about error noticeably.
In general, modeling non-Gaussian event generation for stars of different intensities to improve centroiding remains an open problem, especially for new optical setups where calibration data is not available.
Even without the centroid offset, we achieve similar accuracy results to the original EBS-EKF work. 

The assumption of a perfectly accurate initial estimate of $\ebcincelestial{0}$ also leads to a less accurate estimate of $\ebcinterrestrialreal{}$, as our estimate of the telescope's orientation relative to earth is formed by combining the earth's and telescope's estimated orientations in the celestial frame. 
Estimating $\ebcinterrestrialreal$ remains a difficult problem and introduces additional error on all axes as our slightly incorrect estimate is applied to all IERS earth orientations to form ground truth. 

As previously stated, our telescope mount is unable to hold a static position perfectly over the course of an hour, inducing the obvious drift in estimated declination. 
Improved mounting hardware would reduce this source of error. 

\section{Conclusion}
\label{sec:conclusion}
The accuracy of a recently developed approach to event-based star tracking is quantified in this work using the Earth as a high-accuracy turntable. 
Using the Earth as the source of motion has several advantages: (1) realism, as the star tracker is viewing real (not simulated) stars, and (2) ground truth precision, as the orientation of the Earth is known to a high degree of accuracy.

The benefits of using the Earth's rotation as ground truth are not limited to accuracy quantification experiments. 
While we do not use EBS-EKF's magnitude dependent offset here, modeling the location of the star relative to the positive events it generates is important to accurate tracking.
EBS-EKF notes that the event generation process depends on the image-plane velocity of stars in addition to the magnitude.
The approach presented in this work could be modified to calculate useful ground truth for the image-plane velocity of visible stars, since it provides a mechanism to transfer the high accuracy IERS estimates of the Earth's rotation to the sensor's image plane. 
Future work in this area includes using this image-plane velocity ground truth to develop improved estimation methods for star velocity based on the events being generated.

We show the accuracy of the EBS-EKF approach to star tracking is similar to that of a commercially available star tracker in both across and about accuracy.
Event cameras have the benefits of sparser data streams, higher update rates, lower power usage, and higher dynamic range when compared to framing sensors.
The presented accuracy experiment, combined with these additional benefits, suggests the utility of event cameras for on-orbit star tracking.

{\small
\bibliographystyle{ieeenat_fullname}
\bibliography{11_references}
}


\end{document}


\title{\paperTitle}
\author{\authorBlock}
\maketitlesupplementary

\appendix
\section{Appendix Section}
\label{sec:appendix_section}
Supplementary material goes here.

{\small
\bibliographystyle{ieeenat_fullname}
\bibliography{11_references}
}